\documentclass[10pt,twocolumn,letterpaper]{article}

\usepackage{cvpr}
\usepackage{times}
\usepackage{epsfig}
\usepackage{graphicx}
\usepackage{amsmath}
\usepackage{amssymb}
\usepackage{algorithm}

\usepackage[pagebackref=true,breaklinks=true,letterpaper=true,colorlinks,bookmarks=false]{hyperref}

 \cvprfinalcopy 

\def\cvprPaperID{1572} 

\ifcvprfinal\fi
\begin{document}

\title{Light Field Blind Motion Deblurring}

\author{Pratul P. Srinivasan$^1$, Ren Ng$^1$, Ravi Ramamoorthi$^2$\\
$^1$University of California, Berkeley \qquad $^2$University of California, San Diego\\
{\tt\small $^1$\{pratul,ren\}@eecs.berkeley.edu, $^2$ravir@cs.ucsd.edu}
}

\maketitle
\thispagestyle{empty}

\begin{abstract}

We study the problem of deblurring light fields of general 3D scenes captured under 3D camera motion and present both theoretical and practical contributions. By analyzing the motion-blurred light field in the primal and Fourier domains, we develop intuition into the effects of camera motion on the light field, show the advantages of capturing a 4D light field instead of a conventional 2D image for motion deblurring, and derive simple methods of motion deblurring in certain cases. We then present an algorithm to blindly deblur light fields of general scenes without any estimation of scene geometry, and demonstrate that we can recover both the sharp light field and the 3D camera motion path of real and synthetically-blurred light fields.

\end{abstract}

\vspace{-0.25in}
\section{Introduction}
\vspace{-0.1in}

Motion blur is the result of relative motion between the scene and camera, where photons from a single incoming ray of light are spread over multiple sensor pixels during the exposure. In this work, we make both theoretical and practical contributions by studying the effects of camera motion on light fields and presenting a method to restore motion-blurred light fields. Light field cameras are typically used in situations with optically significant scene depth ranges and out-of-plane camera motion, so it is important to consider how motion blur varies both spatially within each sub-aperture image and angularly between sub-aperture images. 

\begin{figure}
\begin{center}
\newcommand{\width}{1.0\linewidth}
\includegraphics[width=\width]{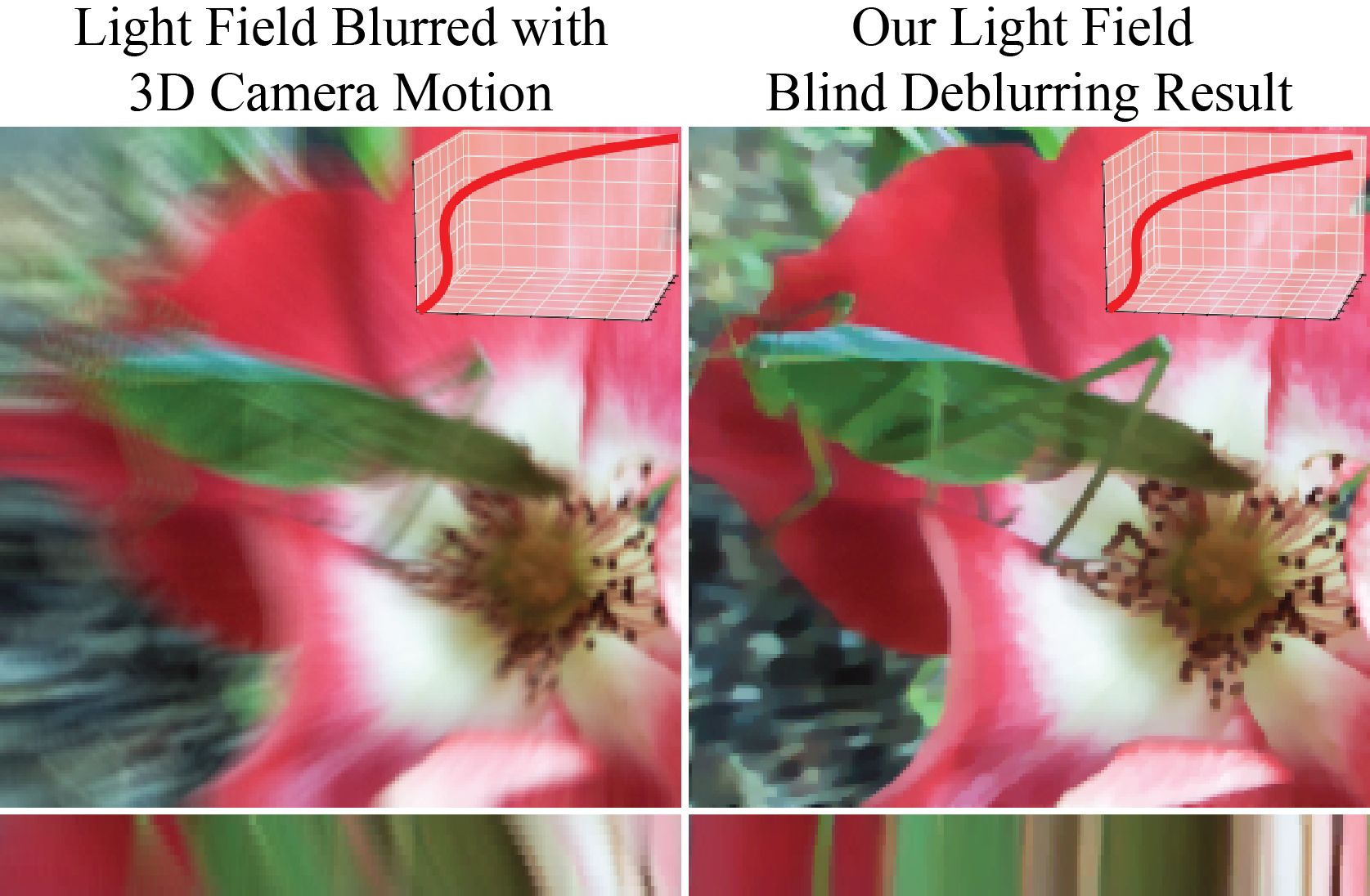} 
\caption{We theoretically study the effects of motion blur on a captured light field and present a practical algorithm to deblur light fields of general scenes captured with 3D camera motion. Left: a 4D light field (visualized as a 2D sub-aperture image and a 2D epipolar slice) is blurred by the synthetic camera motion shown in the inset. Right: absent knowledge of the synthetic motion path, our algorithm is able to accurately recover the sharp light field and the motion path. See Fig.~\ref{fig:real} for examples with real handheld camera motion.}
\label{fig:teaser}
\end{center}
\vspace{-0.32in}
\end{figure}

\vspace{-0.2in}
\paragraph{Theory}

We derive a forward model that describes a motion-blurred light field as an integration over transformations of the sharp light field along the camera motion path. By analyzing the motion-blurred light field in the primal and Fourier domains (Sec.~\ref{sec:theory} and Figs.~\ref{fig:lf_ft_a},~\ref{fig:lf_ft_b},~\ref{fig:lf_ft_c}), we show that capturing a light field enables novel methods of motion deblurring that are not possible with just a conventional image. First, we show that a light field blurred with in-plane camera motion is a simple convolution of the sharp light field with the camera motion path kernel, regardless of the depth contents of the scene (Sec.~\ref{sec:in_plane}). This allows us to use simple deconvolution to restore the sharp light field, which cannot be done with conventional images because the magnitude of the motion blur is depth-dependent (Figs.~\ref{fig:lf_ft_b},~\ref{fig:in_plane_graph}). Additionally, we show that a light field blurred with out-of-plane camera motion is an integral over shears of the sharp light field (Sec.~\ref{sec:out_of_plane}). Therefore, we can blindly deblur a light field of a textured plane captured with out-of-plane camera motion by modulating a slice of the Fourier spectrum of the motion-blurred light field (Figs.~\ref{fig:lf_ft_b},~\ref{fig:out_of_plane_recovery}). This is not possible for conventional images due to the spatially-varying blur caused by out-of-plane camera motion.

\vspace{-0.21in}
\paragraph{Practical Algorithm} 

General light fields of 3D scenes captured with 3D camera motion are integrals over compositions of shears and shifts of the sharp light field. The general light field blind motion deblurring problem lacks a simple analytic approach and is severely ill-posed because there is an infinite set of pairs of sharp light fields and motion paths that explain any observed motion-blurred light field. We propose a practical light field blind motion deblurring algorithm to correct the complex blurring that occurs in situations where light field cameras are useful (Sec.~\ref{sec:algorithm}). Our forward model is differentiable with respect to the camera motion path parameterization and the estimated light field, allowing us to simultaneously solve for both using first-order optimization methods. Furthermore, by treating motion blur as an integration of transformations of the sharp light field, we can simplify the problem by bypassing any estimation of scene geometry. Instead of solving for a dense matrix that represents spatially and angularly varying motion blurs or separately deblurring each sub-aperture image by solving for a 2D blur kernel and 2D depth map, we directly solve for a parameterization of the continuous camera motion curve in $\mathbb{R}^3$. This is a much lower-dimensional optimization problem, and it allows us to utilize the structure of the light field to efficiently recover the motion curve and sharp light field. Finally, we demonstrate the performance of our algorithm on real (Fig.~\ref{fig:real}) and synthetically-blurred (Figs.~\ref{fig:teaser},~\ref{fig:synthetic}) light fields.

\vspace{-0.1in}
\section{Related Work}
\vspace{-0.05in}

\paragraph{Light Fields}

The 4D light field~\cite{Gortler96, Levoy96, Lippmann1908} is the total spatio-angular distribution of light rays passing through free space, and light field cameras capture the light field that exists inside the camera body~\cite{Ng05}. A conventional 2D full-aperture image is produced by integrating the rays entering the entire aperture for each spatial location. Therefore, a captured light field will be interesting and more useful than a conventional image when the equivalent full-aperture image contains significant depth-of-field effects, because this indicates that rays from different regions of the aperture have different values. Common photography situations where capturing a 4D light field would be useful include macro and portrait photography.

Previous work has demonstrated the benefits of lifting problems in computer vision, computer graphics, and computational photography into the 4D light field space. These include rendering 2D pinhole images as slices of the 4D light field~\cite{Levoy96}, stereo reconstruction from a single capture~\cite{Adelson92}, changing the focus and depth-of-field of photographs after capture~\cite{Ng05}, correcting lens aberrations~\cite{Ng06}, passive depth estimation~\cite{Tao15}, glare artifact reduction~\cite{Raskar08}, and scene flow estimation~\cite{Srinivasan15}. 

Previous works have also examined the Fourier spectrum of light fields for various purposes. Chai \etal~\cite{Chai00} analyzed the spectral support of light fields for sampling in light field rendering and showed that Lambertian objects at specific depths correspond to angles in the Fourier domain. Durand \etal~\cite{Durand05} analyzed the effects of shading, occlusion, and propagation on the light field spectrum. Ng~\cite{NgFSP} showed that refocusing a 2D full-aperture image is equivalent to taking 2D slices of the 4D light field spectrum and analyzed the performance of light field refocusing. Liang and Ramamoorthi~\cite{Liang15} developed a light transport framework to investigate the fundamental limits of light field camera resolution. Dansereau \etal~\cite{Dansereau15} derived the 4D spectral support of light fields for rendering, denoising, and refocusing. Additionally, Egan \etal~\cite{Egan09} analyzed the spectrum of motion-blurred 3D space-time images to derive filters for efficient rendering of motion-blurred images. In this work, we analyze the Fourier spectrum of motion-blurred light fields to provide intuition for the effects of camera motion on the captured light field and methods to deblur light fields.

\vspace{-0.2in}
\paragraph{Motion Deblurring}

Blind motion deblurring, removing the motion blur given just a noisy blurred image, is a very challenging problem that has been extensively studied (see~\cite{Lai16} for a recent review and comparison of various algorithms). Representative methods for single image blind deblurring include the variational Bayes approaches of Fergus \etal~\cite{Fergus06} and Levin \etal~\cite{Levin11}, and algorithms using novel image priors such as normalized sparsity~\cite{Krishnan11}, an evolving approximation to the $L_0$ norm~\cite{Xu13}, and $L_0$ norms on both image gradients and intensities~\cite{Pan14}.

Previous multi-image blind deblurring works have also presented algorithms that recover a single 2D image, given multiple observations that have been blurred differently~\cite{Delbracio15, Zhang13, Zhu12}. Jin \etal~\cite{Jin15} present a method that uses a motion-blurred light field of a scene with two depth layers to recover a 2D image and bilayer depth map. Our method also takes a motion-blurred light field as input, but we recover a full 4D deblurred light field as opposed to a 2D texture. Moreover, our method does not need to estimate a depth map.

Many computational photography works have modified the imaging process to make motion deblurring easier. Raskar \etal~\cite{Raskar06} used coded exposures to preserve high frequency details that would be attenuated due to object motion. Another line of work focused on modified imaging methods to engineer point spread functions that would be invariant to object motion. This includes focal sweeps~\cite{Bando13, Kobayashi14}, parabolic camera motions~\cite{Cho10, Levin08}, and circular sensor motions~\cite{Bando11}. In contrast, we focus on the problem of deblurring light fields that have already been captured, and we do not modify the imaging process.

Concurrently with our work, Dansereau \etal~\cite{Dansereau16} introduced a non-blind algorithm to deblur light fields captured with known camera motion.

\vspace{-0.15in}
\section{A Theory of Light Field Motion Blur}
\label{sec:theory}
\vspace{-0.06in}

In our analysis below, we perform a flatland analysis of motion-blurred light fields with a single angular dimension $u$ and a single spatial dimension $x$, and note that it is straightforward to extend this to the full 4D light field with spatial dimensions $(x,y)$ and angular dimensions $(u,v)$. We focus on 3D as opposed to 6D camera motion, so the camera motion path is a general 3D curve and the optical axis does not rotate. 

\vspace{-0.05in}
\subsection{Forward Model}
\vspace{-0.02in}

The observed blurred light field is the integration over the light fields captured at each time $t$ during the exposure:
\vspace{-0.1in}
\begin{equation}
\label{eq:continuous}
\begin{split}
f(x,u) = \int\limits_{t}l_t(x_t,u_t)dt,
\end{split}
\end{equation}
where $f$ is the observed light field and $l_t(x_t,u_t)$ is the sharp light field at time $t$ during the exposure.

\begin{figure}
\begin{center}
\newcommand{\width}{1.0\linewidth}
\includegraphics[width=\width]{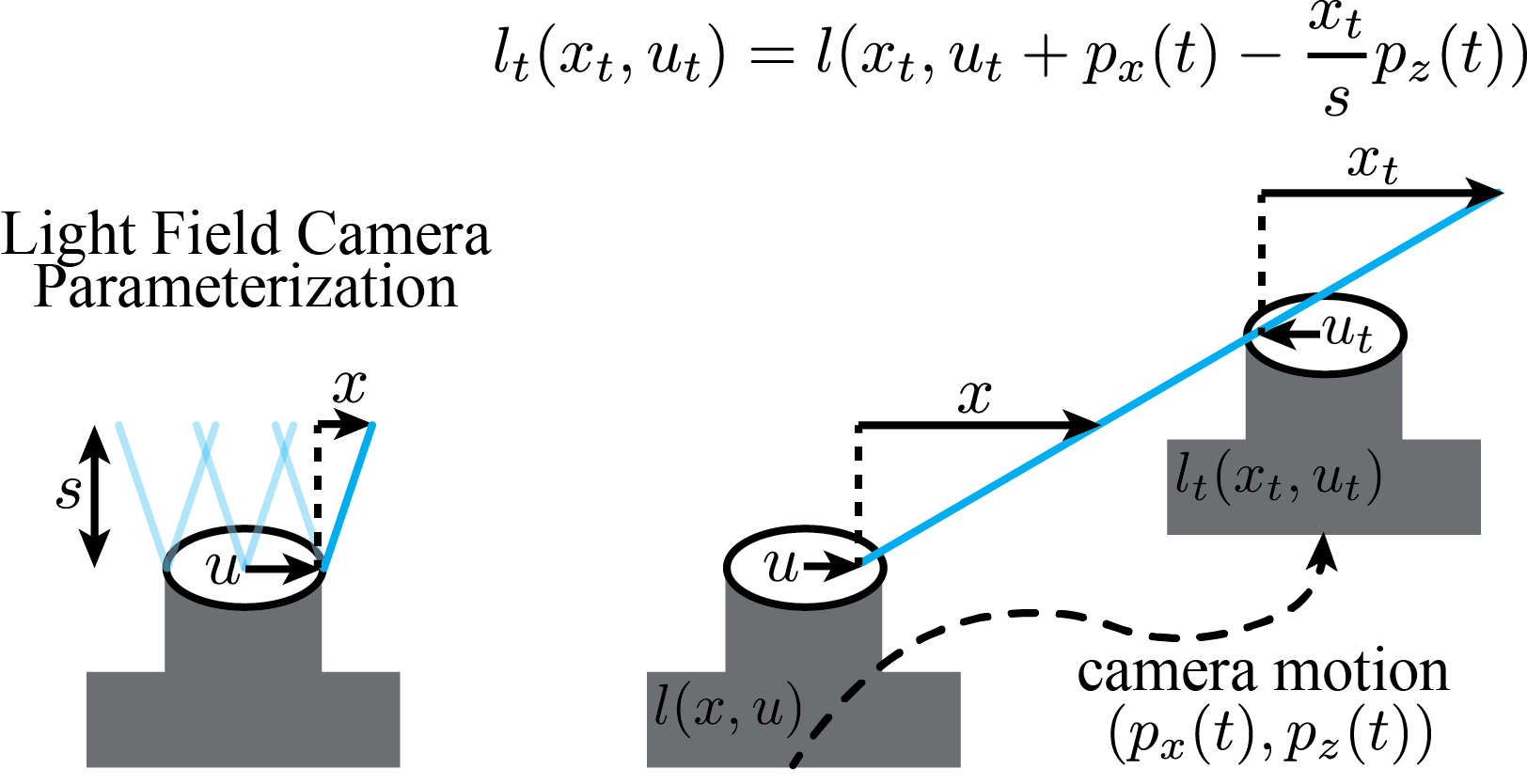} 
\caption{Left: we use a 2-plane parameterization for light fields, where each ray $(x,u)$ is defined by its intercept with the $u$ and $x$ planes separated by distance $s$. Note that the $x$ coordinate is relative to the $u$ coordinate, which is convenient for later derivations. Right: consider a camera translating along a path $\mathbf{p}(t)=(p_x(t),p_y(t),p_z(t))$ during its exposure (in flatland we consider $x$ and $z$ only). The local camera coordinate frame for each time $t$ has its origin located at the center of the camera aperture. The light field $l_t(x_t,u_t)$ is the sharp light field that would have been recorded by the camera at time $t$, in the local camera coordinates at time $t$. The diagram shows that ray $(x_t,u_t)$ in the local coordinate frame at time $t$ is equal to ray $(x_t,u_t+p_x(t)-\frac{x_t}{s}p_z(t))$ in the local coordinate frame at time $t=0$.}
\label{fig:lf_reparam}
\end{center}
\vspace{-0.25in}
\end{figure}

Figure~\ref{fig:lf_reparam} illustrates that the light field at time $t$ is a transformation of the sharp light field at time $t=0$, $l(x,u)$, based on the camera motion path $\mathbf{p}(t)=(p_x(t),p_z(t))$ ($p_y(t)$ is not included in the flatland analysis but is included in the full 3D model). Our light field parameterization is equivalent to considering the light field as a collection of pinhole cameras with centers of projection $u$ and sensor pixels $x$, and we set the separation between the $x$ and $u$ planes $s=1$ so $x$ is a ray's spatial intercept 1 unit above $u$ in the $z$ direction. The observed motion-blurred light field is then

\vspace{-0.1in}
\begin{equation}
\label{eq:forward_continuous}
\begin{split}
f(x,u) = \int\limits_{t}&l(x,u+p_x(t)-xp_z(t))dt.
\end{split}
\end{equation}
\vspace{-0.1in}

Since the light field contains all rays that intersect the two parameterization planes, this forward model accounts for occluded points, as long as the parameterization planes lie outside the convex hull of the visible scene geometry. Certain rare scenarios, such as a macro photography shot where the camera moves between blades of grass during the exposure, may violate this assumption, but it generally holds for typical photography situations. This model also assumes that the light field parameterization planes are infinite, because camera motion can cause the sharp light field at time $t$ to contain rays outside the field-of-view of the light field at a previous time.

\subsection{Space-Angle and Fourier Analysis}

We examine the motion-blurred light field in the primal space-angle and Fourier domains to better understand the effects of camera motion on the captured light field. We denote signals in the Fourier domain with capital letters, and use $\Omega_x$ and $\Omega_u$ to denote spatial and angular frequencies.

It is useful to utilize the Affine Theorem for Fourier transforms~\cite{Bracewell93, Ramamoorthi07}: if $h(\mathbf{a})=g(M\mathbf{b}+\mathbf{c})$, where $M$ is a matrix, $\mathbf{a}$, $\mathbf{b}$, and $\mathbf{c}$ are vectors, and $h$ and $g$ are functions, the relevant Fourier transforms are related as follows: 

\vspace{-0.1in}
\begin{equation}
\label{eq:fourieraffine}
H(\mathbf{\Omega}) = |\mathrm{det}(M)|^{-1} G(M^{-T}\mathbf{\Omega}) \exp(2 \pi i \mathbf{\Omega}^T M^{-1} \mathbf{c}),
\end{equation}
where $\mathrm{det}(M)$ is the determinant of $M$ and $i=\sqrt{-1}$.  

We use this to take the Fourier transform of the observed motion-blurred light field in Eq.~\ref{eq:forward_continuous}, with transformation matrices $M = \left( \begin{smallmatrix} 1&0\\-p_z(t)&1 \end{smallmatrix} \right)$ and $\mathbf{c} = \left( \begin{smallmatrix} 0\\p_x(t) \end{smallmatrix} \right)$:

\vspace{-0.1in}
\small
\begin{equation}
\label{eq:forward_FT_1}
\begin{split}
F(\Omega_x,\Omega_u) = \int\limits_{t} L\left(\Omega_x + p_z(t) \Omega_u,\Omega_u\right)\exp\left[ 2\pi i \Omega_u p_x(t) \right]dt.
\end{split}
\end{equation}
\normalsize
\vspace{-0.1in}

\begin{figure*}[h]
\begin{center}
\newcommand{\width}{1.0\linewidth}
\includegraphics[width=\width]{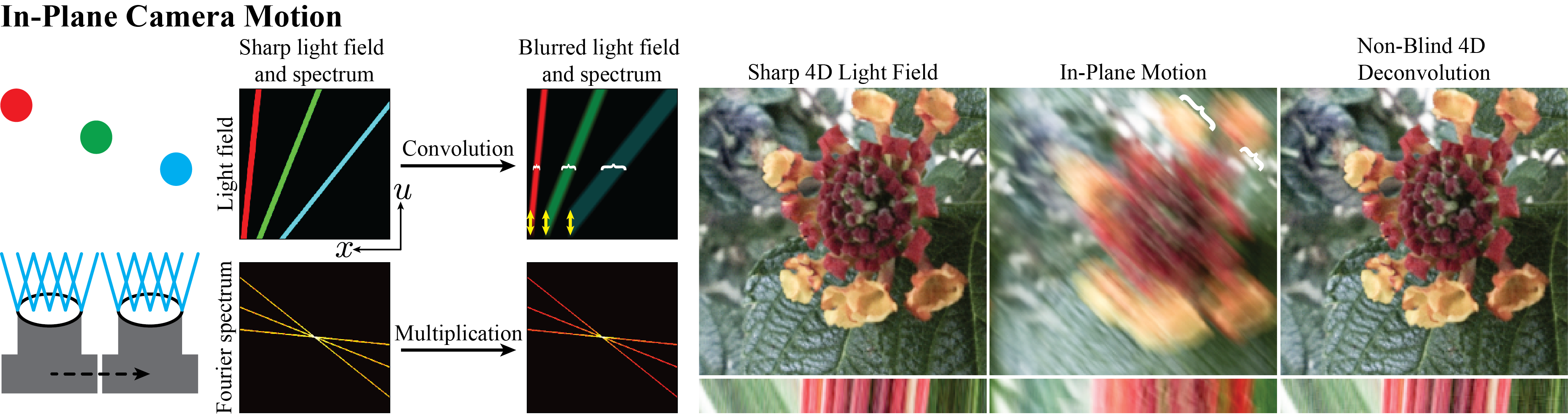} 
\caption{In-plane camera motion is equivalent to a convolution of the light field and the corresponding multiplication of the Fourier spectrum. We are able to easily recover a light field blurred with known in-plane camera motion using 4D deconvolution. Note that in-plane camera motion causes spatially-varying (with $x$) blur due to varying scene depths, as shown by the white brackets, while the blur magnitude does not vary angularly (with $u$), as shown by the yellow vertical arrows.}
\label{fig:lf_ft_a}
\includegraphics[width=\width]{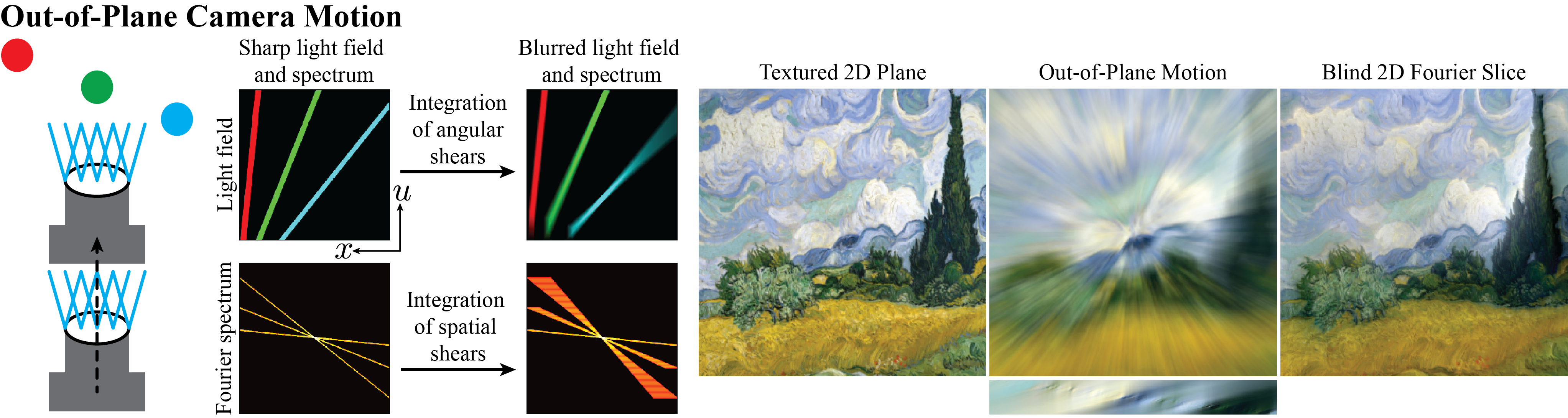} 
\caption{Out-of-plane camera motion is equivalent to an integration over shears in both the primal and Fourier domains. Note that out-of-plane camera motion causes both spatially and angularly varying blur. Given a light field of a single fronto-parallel textured plane (Vincent van Gogh's ``Wheat Field with Cypresses'') with out-of-plane camera motion, we can blindly recover the texture, with slight artifacts due to finite aperture and edge effects, by modulating a 2D slice of the 4D Fourier spectrum.}
\label{fig:lf_ft_b}
\includegraphics[width=\width]{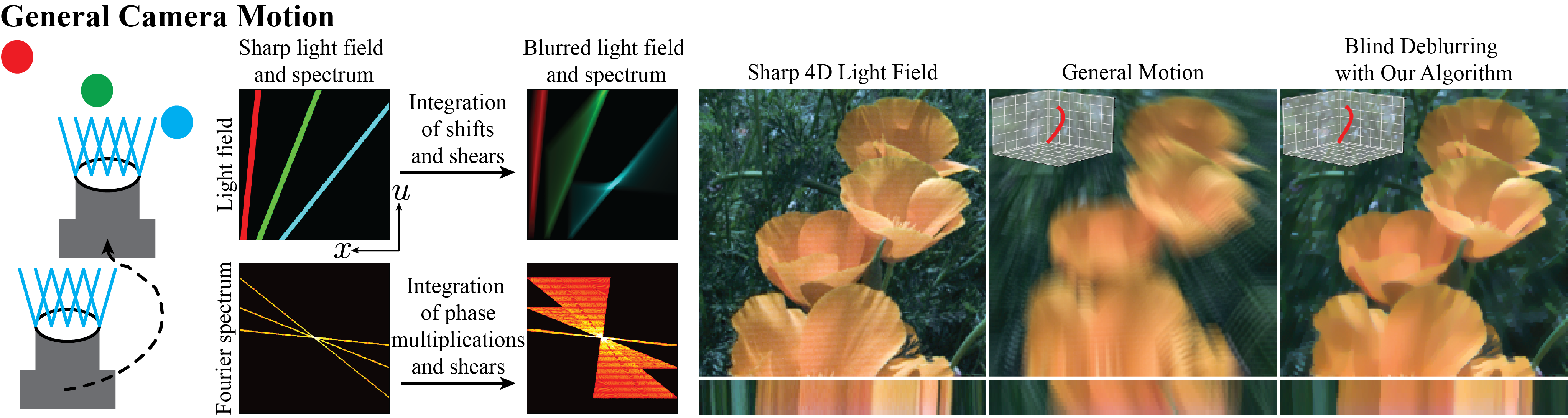} 
\caption{General 3D camera motion is an integration over shears and shifts of the light field and an integration over shears and phase multiplications of the Fourier spectrum. Blindly deblurring light fields captured with general camera motion lacks a simple analytic approach and is severely ill-posed, so we solve this as a regularized inverse problem.}
\label{fig:lf_ft_c}
\end{center}
\vspace{-0.25in}
\end{figure*}

As visualized in Fig.~\ref{fig:lf_ft_c}, the Fourier spectrum is an integration over shears based on the out-of-plane motion $p_z(t)$ and there is also a phase in the complex exponential corresponding to in-plane motion. This complex exponential is the Fourier transform of $\delta(x)\delta(u+p_x(t))$, so we can rewrite the flatland primal domain motion-blurred light field as

\vspace{-0.1in}
\small
\begin{equation}
\label{eq:forward_continuous_conv}
\begin{split}
f(x,u) = \int\limits_{t}[l(x,u-xp_z(t)) \otimes \delta(x)\delta(u+p_x(t))]dt.
\end{split}
\end{equation}
\normalsize
\vspace{-0.1in}

The spatial and frequency domain expressions now separate in-plane motion, which is a convolution with a kernel corresponding to the in-plane camera motion path, and out-of-plane motion, which is an integration over shears in both the spatial and frequency domains. Note that this convolution kernel is restricted to a subspace of the light field space (1D subspace of 2D for flatland light fields, and 2D subspace of 4D for full light fields). 

To gain greater insight into these expressions, we consider two special cases for purely in-plane camera motion, and purely out-of-plane camera motion, with general motion being an integral over compositions of these two cases.

\vspace{-0.1in}
\subsubsection{In-Plane Camera Motion}
\label{sec:in_plane}

For camera motion paths that are parallel to the $x$ and $u$ parameterization planes, $p_z(t)=0$, and the expression for the primal domain motion-blurred light field simplifies to

\vspace{-0.05in}
\begin{equation}
\label{eq:forward_in_plane}
\begin{split}
f(x,u) &= l(x,u) \otimes \int\limits_{t}\delta(x)\delta(u+p_x(t))dt \\&= l(x,u) \otimes \delta(x)k(u),
\end{split}
\end{equation}
where $k(u)=\int\limits_{t}\delta(u+p_x(t))dt$ is the integrated in-plane camera motion path.

In the Fourier domain,

\vspace{-0.1in}
\begin{equation}
\label{eq:forward_in_plane_FT}
\begin{split}
F(\Omega_x,\Omega_u) = &L(\Omega_x,\Omega_u) \int\limits_{t} \exp[ 2\pi i \Omega_u p_x(t) ]dt
\\=&L(\Omega_x,\Omega_u)K(\Omega_u),
\end{split}
\end{equation}
where $K(\Omega_u)=\int\limits_{t} \exp[ 2\pi i \Omega_u p_x(t)]dt$ is the integrated in-plane blur kernel spectrum.

\begin{figure}
\begin{center}
\newcommand{\width}{1.0\linewidth}
\includegraphics[width=\width]{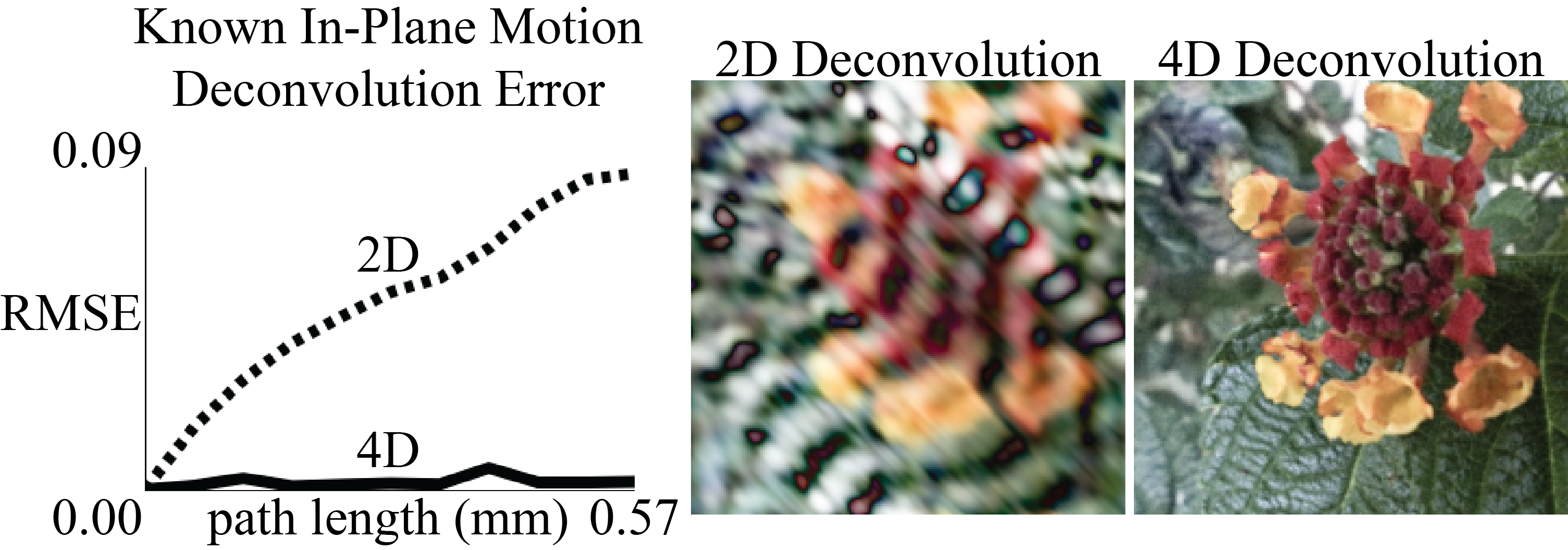} 
\caption{Light fields of general 3D scenes blurred with known in-plane camera motion can be recovered by simple 4D deconvolution. This is not possible with conventional 2D images because the motion blur magnitude is depth-dependent. We synthetically blur a light field with increasing linear in-plane motion, and note that the root mean square error (RMSE) of the central sub-aperture image obtained by 2D deconvolution consistently increases, while the RMSE of the central sub-aperture image obtained by 4D deconvolution of the full light field stays relatively constant.}
\label{fig:in_plane_graph}
\end{center}
\vspace{-0.25in}
\end{figure}

An important insight is that for in-plane camera motion, it is possible to take the original light field out of the integral. {\em This clearly identifies the motion-blurred light field as a simple convolution of the sharp light field with the in-plane blur kernel, regardless of the content and range of depths present in the scene, as illustrated in Fig.~\ref{fig:lf_ft_a}. No such simple result holds for conventional 2D images, as quantified in Fig.~\ref{fig:in_plane_graph}, because the motion blur magnitude is depth-dependent.} Intuitively, in-plane motion is a convolution of the sharp light field because light field cameras at points along the motion path observe the same set of rays shifted, while conventional cameras at points along the motion path observe disjoint sets of rays. If we know the blur kernel, we can recover the sharp light field by simple deconvolution, as shown in Figs.~\ref{fig:lf_ft_a},~\ref{fig:in_plane_graph}. However, if both the blur kernel and light field are unknown, we need to use priors to estimate the blur kernel and sharp light field, as discussed in Sec.~\ref{sec:algorithm}.

\vspace{-0.05in}
\subsubsection{Out-of-Plane Camera Motion}
\label{sec:out_of_plane}

For purely out-of-plane camera motion, $p_x(t)=0$, and the expression for the primal domain motion-blurred light field simplifies to

\vspace{-0.15in}
\begin{equation}
\label{eq:forward_out_of_plane}
\begin{split}
f(x,u) = \int\limits_{t}&l(x,u-xp_z(t))dt
\end{split}
\end{equation}

In the Fourier domain,

\vspace{-0.15in}
\begin{equation}
\label{eq:forward_out_of_plane_FT}
\begin{split}
F(\Omega_x,\Omega_u) = \int\limits_{t} &L(\Omega_x + p_z(t) \Omega_u,\Omega_u)dt.
\end{split}
\end{equation}

These are simply integrations over different shears of the light field, as illustrated in Fig.~\ref{fig:lf_ft_b}. It is particularly interesting to consider the light field of a textured fronto-parallel plane $w(x)$ at depth $z'$. The geometry of our light field parameterization indicates that $l(x,u)=w(xz'+u)$. In the primal domain, the out-of-plane motion-blurred light field of this textured plane is

\vspace{-0.1in}
\small
\begin{equation}
\label{eq:forward_out_of_plane_texture}
\begin{split}
f(x,u) = \int\limits_{t}w(x(z'-p_z(t))+u)dt = \int\limits_{t}w(xz(t)+u)dt,
\end{split}
\end{equation}
\normalsize
where we define $z(t)=z'-p_z(t)$. 

Using the Affine Theorem for Fourier transforms with transformation matrices $M = \left( \begin{smallmatrix} z(t)&1\\0&1 \end{smallmatrix} \right)$ and $\mathbf{c} = \left( \begin{smallmatrix} 0\\0 \end{smallmatrix} \right)$, after noting that the original Fourier transform of the textured plane is $W(\Omega_x)\delta(\Omega_u)$, the Fourier transform of the out-of-plane motion-blurred light field is

\vspace{-0.1in}
\begin{equation}
\label{eq:forward_out_of_plane_texture_FT}
\begin{split}
F(\Omega_x,\Omega_u) = \int\limits_{t} \frac{1}{|z(t)|}
W\left(\frac{\Omega_x}{z(t)}\right) \delta\left(\Omega_u -
  \frac{\Omega_x}{z(t)}\right)\,dt.  
\end{split}
\end{equation}
\vspace{-0.15in}

This is also an integration over various shears, each a line with slope given by $\Omega_u = \Omega_x/z(t)$. The motion-blurred light field takes the original texture frequencies (in a line) and shears them to lines of different slopes, followed by integration. Using the sifting property of the delta function, we can simplify the above expression,

\vspace{-0.1in}
\begin{equation}
\label{eq:forward_out_of_plane_texture_FT_2}
\begin{split}
F(\Omega_x,\Omega_u) = &W(\Omega_u) \int_{z_{\min}}^{z_{\max}} \delta\left(\Omega_u - \frac{\Omega_x}{z}\right) \gamma(z)\,dz, 
\end{split}
\end{equation}
where we have switched to integration over $z$ directly (effectively substituting $z$ for $t$), and for simplicity, we assume $z(t)$ monotonically increases with time. The term $\gamma(z) = (|z| dz/dt)^{-1}$ accounts for the $1/|z|$ factor and change of variables.

Intuitively, the motion-blurred light-field spectrum is a double wedge~\cite{Chai00, Dansereau15, Durand05}, bounded by slopes $z_{\min}$ and $z_{\max}$ and containing an infinite number of lines in the frequency-domain. The magnitudes along each line are the same, determined by the original texture $W(\Omega_u)$, but every value is uniformly scaled by a factor $\gamma(z)$, based on the amount of time the camera lingered at that depth (other than $W(0)$, which is constant for all lines). 

The delta function in Eq.~\ref{eq:forward_out_of_plane_texture_FT_2} can then be evaluated, setting $z=\Omega_x/\Omega_u$, leading to the simple expression,

\begin{equation}
\label{eq:kernelout}
F(\Omega_x,\Omega_u) = W(\Omega_u)\beta\left(\frac{\Omega_x}{\Omega_u}\right),
\end{equation}
where the function $\beta$ includes $\gamma$, as well as the change of variables from the delta function, and is given by 

\vspace{-0.1in}
\begin{equation}
\begin{split}
\beta(z) = \frac{|z|}{\Omega_x dz/dt} \quad \ \beta(\Omega_x/\Omega_u) = \left(|\Omega_u| \left. \frac{dz}{dt}\right|_{\Omega_x/\Omega_u}\right)^{-1}. 
\end{split} 
\end{equation}

\vspace{-0.2in}
\paragraph{Texture Recovery} The structure of the out-of-plane motion-blurred light field enables blind deblurring by a very simple factorization (essentially a rank-1 decomposition of the 2D light field matrix into 1D factors for $W$ and $\gamma$ or $\beta$). One simple approach is to estimate $W$ from any line, then fix the scaling by comparing the overall magnitude of $W$ across lines to estimate the motion blur kernel ($\beta$ or $\gamma$), and finally divide $W(0)$ by the total exposure time.

Taking a slice of the light field in the Fourier domain can be implemented in the primal domain by a sheared integral projection, and this is equivalent to refocusing the full-aperture image to a specific depth~\cite{NgFSP}. Intuitively, this means that blind deblurring of the texture can be performed in the primal domain by computing the full-aperture image refocused to a single depth during the exposure. In summary, {\em we can separately estimate the blur kernel and the original texture for out-of-plane motion of a light field camera, assuming a single fronto-parallel textured plane, by extracting a slice in the Fourier domain or equivalently refocusing the full-aperture image in the primal domain.}

\begin{figure}
\begin{center}
\newcommand{\width}{1.0\linewidth}
\includegraphics[width=\width]{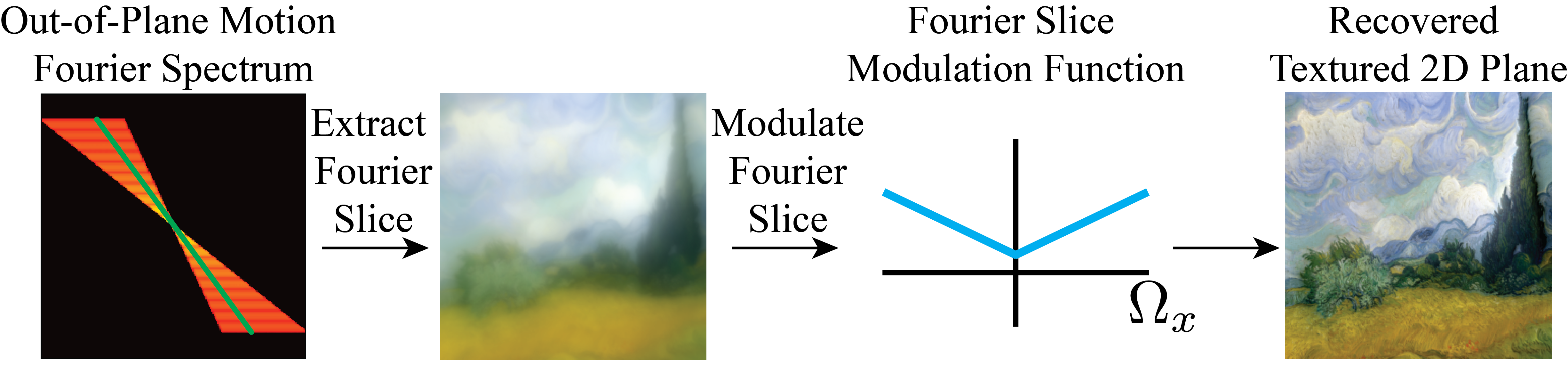} 
\caption{Visualization of process to blindly recover a textured plane from a light field captured with out-of-plane camera motion.}
\label{fig:out_of_plane_recovery}
\end{center}
\vspace{-0.2in}
\end{figure}

Figure~\ref{fig:lf_ft_b} shows an example of a light field of a textured plane blurred by linear out-of-plane motion. We are able to blindly recover the texture, with slight artifacts due to finite aperture and edge effects, by computing a sheared integral projection (equivalent to taking a 2D slice of the 4D Fourier spectrum). As a practical note, when computing this in the discrete case, we must linearly scale frequencies in the extracted slice by $|\zeta\Omega_x|+1$ to correct for the value of each discrete frequency being spread across the shear length during the exposure, where $\zeta$ is a constant corresponding to the relative time the camera lingers at the depth corresponding to that slice. $\zeta$ can be automatically determined by sampling Fourier slices and comparing their magnitudes to calculate the relative time spent at each depth along the motion path. This process is visualized in Fig.~\ref{fig:out_of_plane_recovery}. As detailed in~\cite{NgFSP}, the resolution of the recovered Fourier slice is limited by the angular resolution of the light field camera. 

\vspace{-0.15in}
\paragraph{Comparison with a Conventional Image} It is also insightful to compare this to information available from a conventional 2D image (1D in flatland), corresponding to the view from the central pinhole of a light field camera. In this case, we set $u=0$ in Eq.~\ref{eq:forward_out_of_plane_texture}, defining $l(x) = w(xz')$. Since we are now working in 1D, from the Fourier scale theorem, 
\vspace{-0.1in}
\small
\begin{equation}
F(\Omega_x) = \int\limits_{t} \frac{1}{|z(t)|}
W\left(\frac{\Omega_x}{z(t)}\right)dt=\int_{z_{\min}}^{z_{\max}} 
W\left(\frac{\Omega_x}{z}\right) \gamma(z)dz.
\end{equation}
\normalsize

This is similar to the light field case, except that we no longer have the delta function for multiple sheared lines in 2D; indeed we only have a single 1D line, with a frequency spectrum scaled according to $z$. It is clear that from the perspective of analysis and recovery, the conventional image case provides far less insight than in the light field case. We cannot separate out the texture $W$, and the methods for recovery discussed in the light field case do not apply, since there are not multiple lines in a 2D spectrum we can study. In fact, it is not even straightforward to recover texture and motion blur kernel even when one of the factors is known.

\vspace{-0.15in}
\subsubsection{General 3D Motion and Scenes}
\vspace{-0.05in}

General 3D camera motion is an integral over compositions of shears and shifts of the light field, as shown in Fig.~\ref{fig:lf_ft_c}. Blindly deblurring a light field of a general scene captured with 3D camera motion lacks a simple analytic approach and is a severely ill-posed problem because there is an infinite set of pairs of light fields and motion paths that explain any observed motion-blurred light field. Below, we present an algorithm to estimate the sharp light field and camera motion path by solving a regularized inverse problem.

\begin{figure*}
\begin{center}
\newcommand{\width}{1.0\linewidth}
\includegraphics[width=\width]{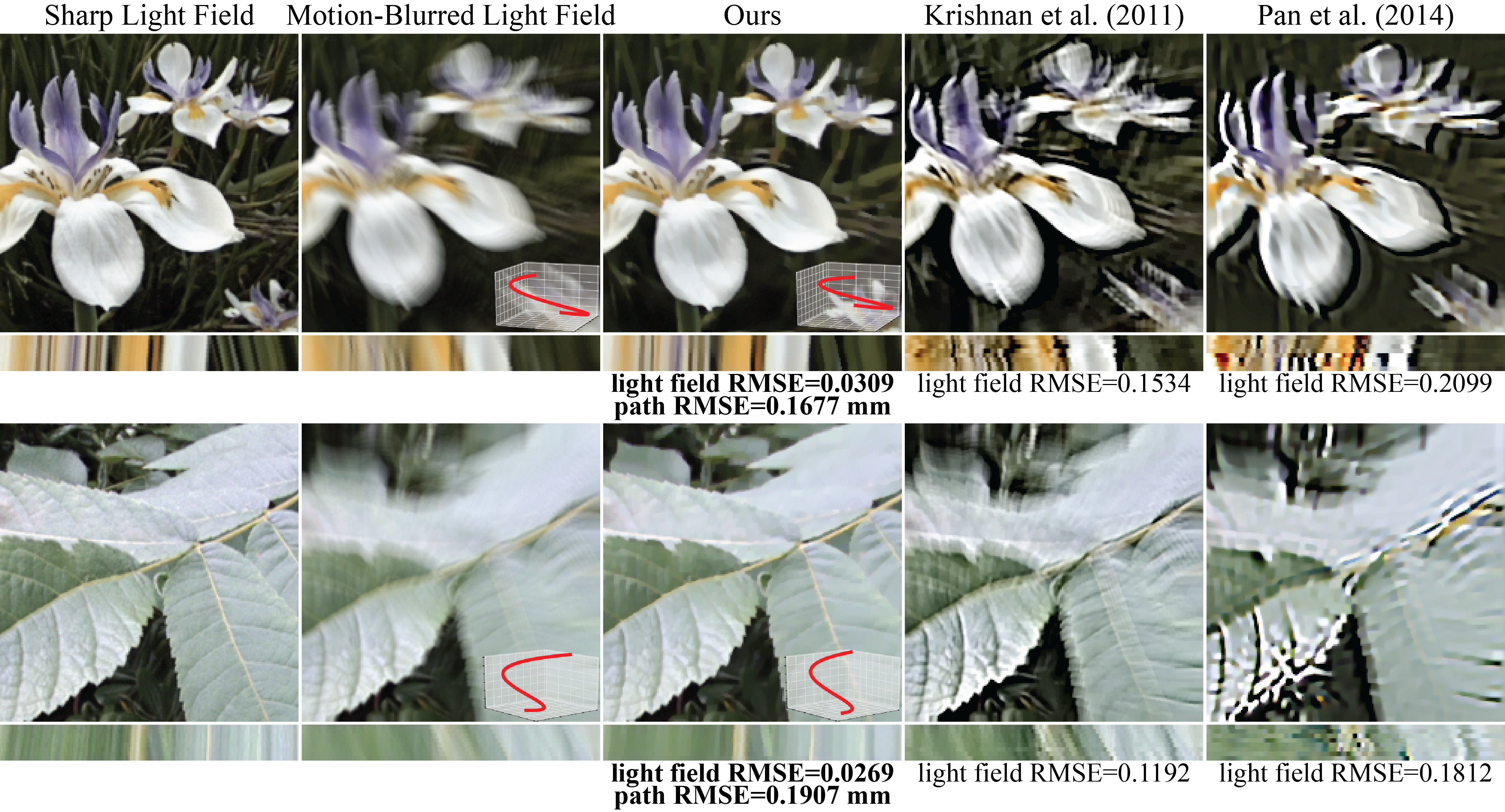} 
\caption{Blind deblurring results on synthetically motion-blurred light fields. Our algorithm is able to correctly recover the sharp light field and estimate the 3D camera motion path, while alternative methods perform poorly due to the large spatial variance in the blur. Additionally, as demonstrated by the epipolar images, other algorithms do not recover a light field that is consistent across angular dimensions. The root mean square error (RMSE) of our deblurred results are consistently lower than those of the alternative methods.}
\label{fig:synthetic}
\end{center}
\vspace{-0.2in}
\end{figure*}

\vspace{-0.1in}
\section{Blind Light Field Deblurring Algorithm}
\label{sec:algorithm}
\vspace{-0.05in}

For blind light field motion deblurring, we estimate both the camera motion curve $\mathbf{p}(t)$ and the sharp light field $\mathbf{l}$. We utilize our forward model derived in Eq.~\ref{eq:forward_continuous} to formulate a regularized inverse problem, and our approach is particularly efficient due to our direct representation of the camera motion curve, as discussed below. We solve a discrete optimization problem, since light field cameras record samples and not continuous functions:

\vspace{-0.15in}
\begin{equation}
\label{eq:optimization_discrete}
\begin{split}
\underset{\mathbf{l},\mathbf{p}(t)}{{\operatorname{min}}}||\mathbf{\hat{f}}(\mathbf{l},\mathbf{p}(t))-\mathbf{f}||^2_2+\lambda\psi(\mathbf{l}),
\end{split}
\end{equation}
where the first term minimizes the $L_2$ norm of the difference between the observed motion-blurred light field $\mathbf{f}$ and that predicted by the forward model $\mathbf{\hat{f}}$, and the second term, $\psi(\mathbf{l})$, is a prior on the sharp light field. To address finite aperture and sensor planes, we assume replicating boundaries for the sharp light field. We use bilinear interpolation to transform the sharp light field along the camera motion path, so our forward model is differentiable with respect to the camera path and the sharp light field.

\vspace{-0.15in}
\paragraph{Camera Motion Path Representation}

We model the camera motion path $\mathbf{p}(t)$ as a B\'ezier curve made up of $n$ control points in $\mathbb{R}^3$. This approach is much more efficient than the alternative approaches of solving for a dense matrix to represent spatially and angularly varying blurs, or separately deblurring each sub-aperture image. A dense motion blur ray transfer matrix would have size $r \times r$, where $r$ is the number of rays sampled by the light field camera (this matrix would have size $2560000 \times 2560000$ for the light fields used in this work). Separately deblurring each sub-aperture image involves estimating a 2D depth map and a 2D convolution kernel, each of size $s \times s$, where $s$ is the number of samples along each spatial dimension (this equates to solving for two matrices of size $200 \times 200$ for the light fields used in this work). Instead, we solve for a much lower-dimensional vector of control points with $3n$ elements. In practice, we find that typical camera motion paths can be represented by $n=3$ or $n=4$ control points. 

\vspace{-0.1in}
\paragraph{Light Field Prior}

To regularize the inverse problem above, we use a 4D version of the sparse gradient prior proposed in~\cite{Xu13}:

\vspace{-0.15in}
\begin{equation}
\label{eq:lf_prior}
\begin{split}
\psi(\mathbf{l}) = \sum\limits_{x,y,u,v}
\begin{cases}
\frac{1}{\epsilon^2}|\nabla\mathbf{l}|^2 & \text{if } |\nabla\mathbf{l}|\leq\epsilon,\\
1 & \text{otherwise}.\\
\end{cases}
\end{split}
\end{equation}

This function gradually approximates the $L_0$ norm of gradients by thresholding a quadratic penalty function parameterized by $\epsilon$, and approaches the $L_0$ norm as $\epsilon\to0$.

\vspace{-0.15in}
\paragraph{Implementation Details}

We utilize the automatic differentiation of Tensorflow~\cite{Tensorflow2015-whitepaper} to differentiate the loss of the blind deblurring problem in Eq.~\ref{eq:optimization_discrete} with respect to the camera motion path control points and sharp light field, and use the first-order Adam solver~\cite{Kingma15} for optimization.

While the prior in Eq.~\ref{eq:lf_prior} is effective for estimating the camera motion path, the sharp light fields estimated using this prior typically appear unnatural and over-regularized. We hold the camera motion path constant and solve Eq.~\ref{eq:optimization_discrete} for the sharp light field using a 4D total variation ($L_1$ norm of gradients) prior to obtain the final sharp light field.

\begin{figure*}
\vspace{-0.2in}
\begin{center}
\newcommand{\width}{1.0\linewidth}
\includegraphics[width=\width]{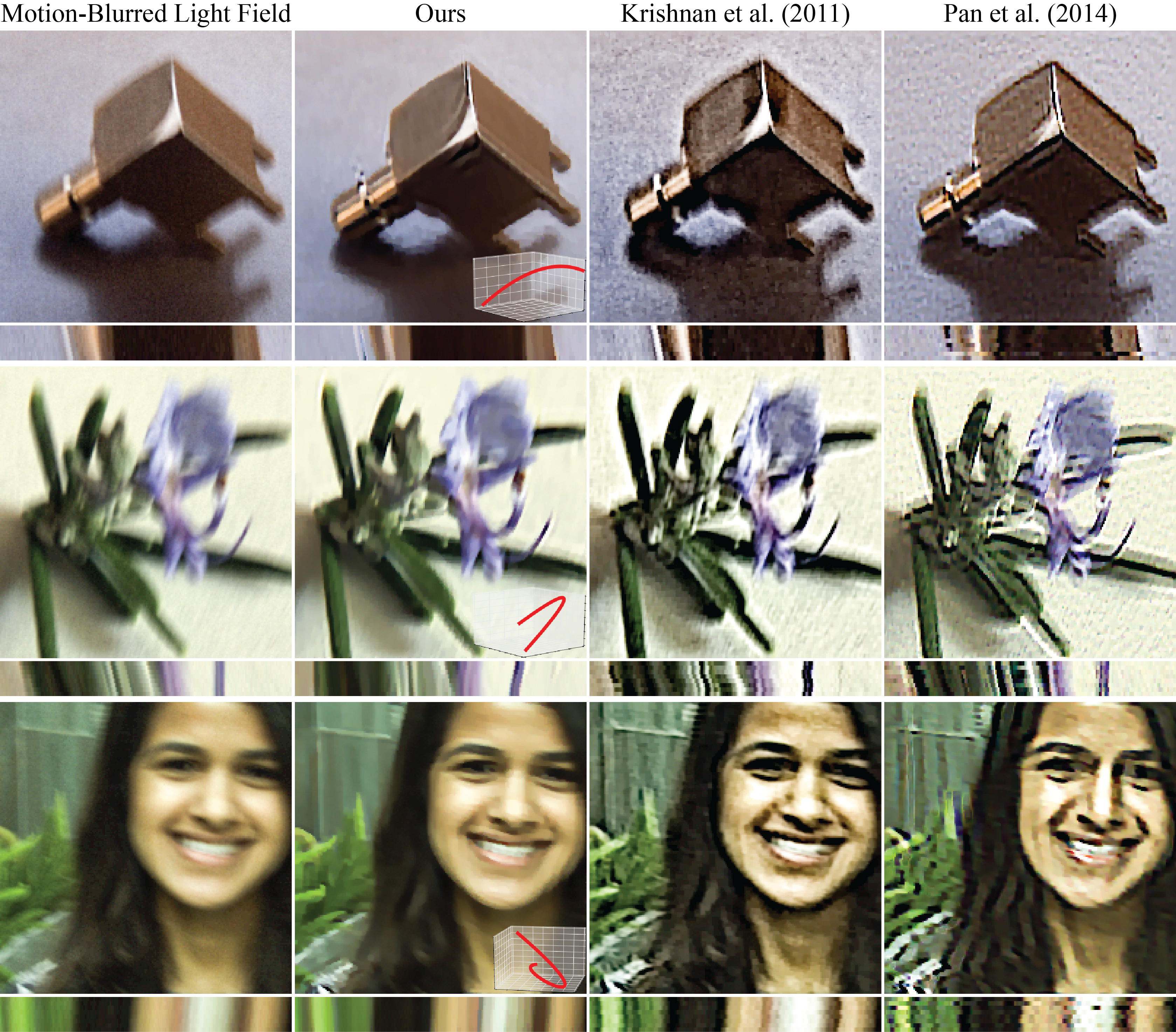} 
\caption{Blind deblurring results on real handheld motion-blurred light fields. Our algorithm is able to correctly recover the sharp light field and estimate the 3D camera motion path. Note that we correct the motion seen in the specular reflections and object edges in the circuit component example, the motion seen in the leaves and flower in the rosemary plant example, and the motion seen in the hair, eyebrows, teeth, and background plants of the portrait example. Furthermore, our method produces angularly-consistent results, as demonstrated by the epipolar slices of all 3 examples. Please view our supplementary video and project webpage for animated visualizations of our results.}
\label{fig:real}
\end{center}
\vspace{-0.3in}
\end{figure*}

\vspace{-0.05in}
\subsection{Results}
\label{sec:results}
\vspace{-0.05in}
We validate our algorithm using light fields captured with the Lytro Illum camera, that have been blurred by both synthetic camera motion within a 2 mm cube using our forward model in Eq.~\ref{eq:forward_continuous} and real handheld camera motion using a shutter speed of 1/20 second. We compare our results to the alternative of applying state-of-the-art blind image motion deblurring algorithms to each sub-aperture image. As shown in a recent review and comparison paper~\cite{Lai16}, the algorithms of Krishnan \etal~\cite{Krishnan11} and Pan \etal~\cite{Pan14} are two of the top performers for blind deblurring of both real and synthetic images with spatially-varying blur, so we compare our algorithm to these two methods. As demonstrated by both the synthetically motion-blurred results in Fig.~\ref{fig:synthetic} and the real motion-blurred results in Fig.~\ref{fig:real}, our algorithm is able to accurately estimate both the sharp light field and the camera motion path. The state-of-the-art blind image motion deblurring algorithms are not as successful due to the significant spatial variance of the blur. Furthermore, they are not designed to take advantage of the light field structure and do not estimate a 3D camera motion path, so their results are inconsistent between sub-aperture images, as demonstrated by the epipolar image results.

In the synthetically-blurred examples in Fig.~\ref{fig:synthetic}, note that our algorithm correctly estimates the ground truth complex camera motion paths and corrects the large spatially-varying blurs in the flowers and leaves. In the real handheld blurred examples in Fig.~\ref{fig:real}, note that our algorithm corrects the blur in the specularities and edges of the circuit component, the leaves and flower of the rosemary plant, and the hair, eyebrows, teeth, and background plants. 

\section{Conclusion}
\vspace{-0.15in}
In this work, we studied the problem of deblurring light fields of general scenes captured with 3D camera motion. We analyzed the effects of motion blur on the light field in the primal and Fourier domains, derived simple methods to deblur light fields in specific cases, and presented an algorithm to infer the sharp light field and camera motion path from real and synthetically-blurred light fields. It would be interesting to extend our forward model to account for 3D rotations of the optical axis, and theoretically analyze the effects of camera rotation on the motion-blurred light field. Since the forward model would be differentiable with respect to the rotation parameters, our blind deblurring optimization algorithm can easily generalize to account for camera rotation.

We think that the insights of this work enable future investigations of light field priors that more explicitly consider the effects of motion blur on the light field, as well as novel interpretations of single and multi-image motion deblurring as subsets of the general light field motion deblurring problem. 

\vspace{-0.25in}
\paragraph{Acknowledgments} This work was supported in part by ONR grant N00014152013, NSF grant 1617234, NSF graduate research fellowship DGE 1106400, a Google Research Award, the UC San Diego Center for Visual Computing, and a GPU donation from NVIDIA.

{\small
\bibliographystyle{ieee}
\bibliography{PRATUL}
}

\end{document}